\documentclass{article}

\usepackage{microtype}
\usepackage{graphicx}
\usepackage{subcaption}
\usepackage{booktabs} 

\usepackage{hyperref}

\usepackage[accepted]{icml2026}

\usepackage{amsmath}
\usepackage{amssymb}
\usepackage{mathtools}
\usepackage{amsthm}

\usepackage[capitalize,noabbrev]{cleveref}

\theoremstyle{plain}
\newtheorem{theorem}{Theorem}[section]

\theoremstyle{definition}
\newtheorem{definition}[theorem]{Definition}

\theoremstyle{remark}

\usepackage[textsize=tiny]{todonotes}

\usepackage{xcolor}
\usepackage{enumitem}

\newcommand{\prob}{\mathbb{P}}
\newcommand{\score}{\hat{\mu}}
\newcommand{\truescore}{\mu}

\usepackage{layouts}
\icmltitlerunning{Quantifying Ranking Uncertainty in LLM Benchmarks}

\begin{document}

\twocolumn[
  \icmltitle{Quantifying Ranking Uncertainty in LLM Benchmarks}



  \icmlsetsymbol{equal}{*}

  \begin{icmlauthorlist}
    \icmlauthor{Bitya Neuhof}{yyy}
    \icmlauthor{Yuval Benjamini}{yyy}
  \end{icmlauthorlist}

  \icmlaffiliation{yyy}{Department of Statistics and Data Science, The Hebrew University of Jerusalem, Jerusalem, Israel}

  \icmlcorrespondingauthor{Bitya Neuhof}{bitya.neuhof@mail.huji.ac.il}
  \icmlcorrespondingauthor{Yuval Benjamini}{yuval.benjamini@mail.huji.ac.il}

  \icmlkeywords{Machine Learning, ICML, AI, leaderboards, ranking}

  \vskip 0.3in
]



\printAffiliationsAndNotice{}  

\begin{abstract}
    Pretrained models are typically ranked on multi-task leaderboards to assess their effectiveness across diverse tasks. Rank confidence intervals were recently introduced as a method to quantify the uncertainty in these rankings by aggregating pairwise hypothesis tests. In this work, we analyze the sources of uncertainty in the knowledge evaluation benchmark MMLU and show how hypothesis tests can be modified to account for their effects. We demonstrate that ranking variability across MMLU subjects is substantial and should be considered when comparing LLMs or identifying the top-performing models.
    \end{abstract}

\section{Introduction}

The practice of model ranking has been formalized through public leaderboards, such as the Hugging Face Open LLM Leaderboard~\cite{open-llm-leaderboard-v2}, which provides an overview of model rankings across diverse tasks. As a result, leaderboards have become the primary way researchers report empirical progress, with rankings signaling advancement and fostering competition. For each benchmark, models are evaluated on multiple tasks and prompts. Because evaluation metrics use different numerical scales~\cite{demvsar2006statistical, longjohn2025statistical}, a summary performance metric is typically ordered and presented as a ranking. These rankings then guide model comparison and selection. Despite the widespread adoption of rankings, important limitations remain. Leaderboard results can vary significantly depending on the evaluation set and the consistency of evaluation methods~\cite{rising2021uncertainty}. Moreover, rankings are typically presented as single values, without any quantification of their uncertainty~\cite{miller2024adding, ackerman2025statistical}. This lack of uncertainty quantification can obscure substantial variation in model performance across tasks or prompts, especially for large, general-purpose pretrained models (such as LLMs). While confidence intervals (CIs) for model scores are common,  CIs for their ranks are needed to support the identification of significant model differences~\cite{foucart2025ranking, valdeira2025ranking}.

In this paper, we discuss how directional pairwise tests and rank CIs can improve uncertainty estimation for ranking. We demonstrate the role of tests on the Massive Multitask Language Understanding (MMLU) benchmark~\cite{hendryckstest2021, polo2024efficient}, which consists of multiple-choice questions across a wide range of subjects. Our focus is on the choice of hypothesis tests underlying the construction of the rank CIs on the resulting intervals. Using these tools, we identify sources of ranking uncertainty, show that variability between subjects exceeds that between prompt variants, and highlight the importance of subject-level analysis for detecting performance differences between models.
The code for our analyses is publicly available at \href{https://github.com/BityaNeuhof/quantifying-rank-uncertainty.git}{quantifying-rank-uncertainty}.

\section{Rank Confidence Intervals}\label{sec:rank_ci}

In this section, we describe a method for constructing rank confidence intervals (CIs) from paired samples~\cite{holm2013confidence, al2021simultaneous, neuhof2024confident, foucart2025ranking}. Rank CIs provide a principled way to communicate the uncertainty in model rankings, which is not captured by reporting only raw ranks. Here, we treat the sources of uncertainty or levels of aggregation as given. In the next section, we examine the influence this choice has on how the results are interpreted.

\subsection{Terminology and Definitions}

Let $\mathcal{M}$ be a set of models, with $N_{\mathcal{M}}$ models. For simplicity, assume a balanced design, meaning that for each model $m_j$ we observe a vector of per-unit performance scores $s_{ij}$ for units $y_i$, $i=1, \ldots ,n$. Higher scores mean better performance. The observed model scores $\score = (\score_1,...,\score_{N_{\mathcal{M}}})$ are obtained by averaging over the units $\score_j = \frac{1}{n}\sum s_{ij}$. Raw observed ranks $\mathbf{\hat{r}} = (\hat{r}_1, \ldots, \hat{r}_{N_{\mathcal{M}}})$, with $\hat{r}_j \in \{1, \ldots, {N_{\mathcal{M}}}\}$, are assigned by ordering $\mathbf{\score}$ so the highest score gets rank ${N_{\mathcal{M}}}$ and the lowest gets rank $1$.

Let $\truescore_j$ denote the true performance score of model $m_j$, and $\score_j$ denote its noisy estimate. The true rank of model $m_j$ is determined by the relative position of $\truescore_j$ among the true scores $(\truescore_1, \ldots, \truescore_{N_{\mathcal{M}}})$. When there are no ties, the true rank of $m_j$ can be written in two equivalent ways:
\begin{equation}\label{eq:true_rank_two_forms}
    \begin{aligned}
    r_j &= 1 + \#\{\truescore_k > \truescore_j :k \neq j\} \\
    & = N_{\mathcal{M}} - \#\{\truescore_k < \truescore_j: k \neq j\}
    \end{aligned}
\end{equation}
The first expression counts how many models outperform $m_j$; the second counts how many are outperformed by $m_j$. This equivalence enables the construction of rank CIs using directional pairwise comparisons: instead of the exact true scores, we rely on statistical evidence to identify models that are significantly better or worse than $m_j$. These identified groups provide upper and lower bounds on the possible rank of $m_j$. Rank intervals can be used to communicate the uncertainty in the true ranks. Let $[L_j, U_j]$ be the rank CI for model $m_j$, with $\alpha$ as the miscoverage rate. Here, $r_j$ denotes the true rank of model $m_j$. If there are ties in the true scores, $r_j$ may be a set of possible ranks rather than a single value. In this case, the coverage definition below treats $r_j$ as a set and checks whether it is contained in the interval.
\begin{definition}(Valid Rank CI)
    \label{def:valid_ci}
    A rank interval for model $m_j$ is valid if the probability that the true rank $r_j$ is not contained in the interval $[L_j, U_j]$ is at most $\alpha$.
    \begin{equation}\label{eq:valid_ci}
        \begin{aligned}
        \prob \big( r_j \subseteq [L_j, U_j]\big) \geq 1 - \alpha.
        \end{aligned}
    \end{equation}
\end{definition}
        
A set of valid rank CIs may guarantee coverage for each model individually (marginal coverage) or for all models simultaneously (simultaneous coverage). See Appendix~\ref{app:rank_ci} for formal definitions.

We construct rank CIs using pairwise location tests, extending the standard approach of sorting raw scores. This method allows control over ranking errors, either at the individual model level or across all models~\cite{holm2013confidence, al2022simultaneous}.
For each pair of models, with true scores $\truescore_j$ and $\truescore_k$, we test two one-sided hypotheses:
\begin{equation}\label{eq:hypotheses}
    \begin{aligned}
        &H_{jk;0}: \truescore_j \leq \truescore_k \text{ vs } H_{jk;1}: \truescore_j > \truescore_k, \text{ and } \\
        &H_{kj;0}: \truescore_k \leq \truescore_j \text{ vs } H_{kj;1}: \truescore_k > \truescore_j.
    \end{aligned}
\end{equation}
The result of each hypothesis test is a p-value $p_{jk}$ for every ordered pair of models $(m_j, m_k)$.

For $N_{\mathcal{M}}$ models, there are $N_{\mathcal{M}}(N_{\mathcal{M}}-1)$ such pairwise comparisons. Rank CIs for the true ranks are obtained by \emph{counting} rejections of hypotheses after controlling for family-wise error rate (FWER). FWER control can be performed for decisions associated with a single model at a time~\cite{holm2013confidence, Mogstad2024} or simultaneously across all models~\cite{al2022simultaneous, neuhof2024confident}.

Given a set of hypotheses $\{H_{jk;0}, j \neq k\}$, the lower and upper bounds of the ranks are:
\begin{equation}\label{eq:rank_ci}
    \begin{aligned}
        &L_j =1 + \#\{k \neq j : H_{jk;0} \text{ is rejected}\}, \\
        &U_j =M - \#\{k \neq j : H_{kj;0} \text{ is rejected}\}.
    \end{aligned}
\end{equation}
Thus, the lower and upper rank bounds are obtained by applying the same logic in opposite directions: significant evidence that other models are better moves $U_j$ downward to a worse (lower) rank, while significant evidence that other models are worse moves $L_j$ to a better (higher) rank. In our implementation, we control for multiple comparisons using Holm’s correction~\cite{holm1979simple}.
The algorithms to construct rank CIs from the pairwise comparisons, as well as proofs of validity, are detailed in Appendix~\ref{app:rank_ci}.

\subsection{Statistical Choices in Balanced Designs}

The construction of rank CIs from pairwise location tests leaves central statistical questions: which hypotheses to test, which population to consider, and which error to control. These choices correspond to distinct probability spaces and yield distinct interpretations of the resulting rank intervals. Importantly, these abstract decisions directly determine the statistical procedures and the interpretation of the results; for example, the choice of population affects which units are compared, while the error-control target shapes the confidence guarantees. The rest of the paper uses this distinction to clarify which kind of ranking uncertainty is being quantified and why the choice of hypothesis-testing procedure and error-control target is not merely a technical detail but an integral part of the inferential claim being made. We demonstrate the different forms of uncertainty and how they translate to pairwise comparisons through the paired t-test. In a balanced design with approximately normally distributed data, define the p-value $p_{jk}$ as $p_{jk} = T_{n-1}(t)$, where $t = \frac{\bar{d}}{\sqrt{n}sd(\bar{d})}$. Here, $\bar{d}$ is the difference of the sample means, $n$ is the sample size, and $sd(\bar{d})$ is the sample standard deviation (SD) of the difference. $T_{n-1}$ denotes the cumulative distribution function (CDF) of the T-distribution with $n-1$ degrees of freedom. Robust alternatives use a variant of $t$ and estimate its distribution from the sample, for example, using bootstrap or permutation methods~\cite{wilcox2023comparing}; more details and an example are in Appendix~\ref{app:rank_ci} and \ref{fig:app_rank_ci_wilcoxon}. We use $n$ for the number of units being compared, but $n$ varies across definitions of units. In unbalanced designs, a similar approach can also be applied with a random-effects or mixed-effects model~\cite{JSSv067i01}.
\section{Rank CIs Application to MMLU}\label{sec:mmlu}

The MMLU benchmark~\cite{hendryckstest2021} evaluates LLMs' knowledge and problem-solving abilities across a wide range of subjects, spanning STEM, humanities, and social sciences, via multiple-choice questions. We use the PromptEval extension~\cite{polo2024efficient} of MMLU, which provides a correctness matrix covering 100 prompt variations, 15 models, and 57 subjects, each with at least 100 questions. The data is publicly available (MIT license) to download from Hugging Face~\footnote {\url{https://huggingface.co/datasets/PromptEval/PromptEval_MMLU_correctness}}. PromptEval was designed to evaluate model sensitivity to prompt variation, introducing two main sources of variability for each subject: across prompts and across questions. In addition, we expect variability between subjects and model correlations that may differ by subject. In a complex benchmark like MMLU, relying on a single average accuracy across subjects risks obscuring both between-subject and within-subject variability. Such variability arises due to differences in subject domains, question content, and the number of questions per subject. To make subjects comparable, we randomly sample 100 questions from each subject.

We measure variability by constructing rank CIs for models, using the hypothesis tests described in Section~\ref{sec:rank_ci} (Equation~\ref{eq:hypotheses}). Let $\mathcal{M}$, $\mathcal{T}$, $\mathcal{Q}(\mathcal{T})$, and $\mathcal{V}$ denote the sets of models, subjects, questions, and prompt variants, respectively. Define $S: (\mathcal{M}, \mathcal{T}, \mathcal{Q}(\mathcal{T}), \mathcal{V}) \to \{0, 1\}$ as a binary function that assigns a score of 1 for correct answers and 0 for incorrect ones. Let $s_{ij}$ denote the score of model $m_j$ and unit $y_i$, where each unit may correspond, for example, to a single question-prompt pair for a given subject, or to an aggregation over multiple questions and prompts, depending on the analysis context. This definition of a unit forms the basis for our hypothesis tests and for constructing the vector of differences between the models analyzed in this paper. The total number of units, $n$, constitutes the population for the paired t-test. All reported score and rank CIs use a miscoverage level of $\alpha = 0.05$. Note that all the rank CIs in this section have marginal coverage; an example of simultaneous rank CIs is provided in Appendix~\ref{app:add_simultaneous}.

\subsection{Benchmark-Level Variability}

A naive way to compare two models is to treat each individual observation as a separate unit. Specifically, a unit $y_i$ is indexed by all unique triples $(t_b, q_{bc}, v_l)$, where $t_b$ is a subject, $q_{bc}$ is a question for subject $t_b$, and $v_l$ is a prompt variant. 
For a pair of models $(m_j, m_k)$, we form:
\begin{equation}\label{eq:global_diff}
\begin{aligned}
    d_{[jk,b,bc,l]} = S(m_j,t_b,q_{bc},v_l) - S(m_k,t_b,q_{bc},v_l),
\end{aligned}
\end{equation}
and test whether the average of these unit-level differences is zero.
For models $(m_j, m_k)$, the estimator is
\begin{equation}\label{eq:global_mu}
\begin{aligned}
        \bar{d} = \frac{1}{N_{\mathcal{T}}N_{\mathcal{Q}(\mathcal{T})}N_{\mathcal{V}}}
        \sum_b \sum_c \sum_l d_{[jk,b,bc,l]} = \score_j - \score_k
\end{aligned}
\end{equation}
where $N_{\mathcal{T}}$ is the number of subjects, $N_{\mathcal{Q}(\mathcal{T})}$ is the number of questions per subject, and $N_{\mathcal{V}}$ is the number of prompt variants. The sample size is therefore the number of units $n=N_{\mathcal{T}}N_{\mathcal{Q}(\mathcal{T})}N_{\mathcal{V}}$, and the SD measures the spread of $d_{[jk,b,bc,l]}$ around $\bar{d}$.

\begin{figure}[ht]
    \centering
    \includegraphics[width=0.98\linewidth]{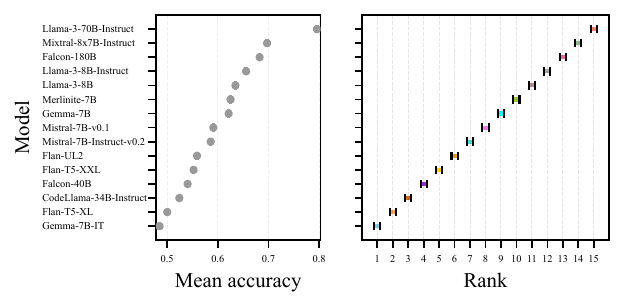}
    \caption{Scores with 95\% CIs and rank CIs across all subjects, prompt variants, and questions. The models are clearly distinguishable, as indicated by non-overlapping CIs.}
    \label{fig:global_ranking_and_score}
\end{figure}

In Figure~\ref{fig:global_ranking_and_score} we show the CIs for the scores, alongside the rank CIs obtained from the differences described by Equation~\ref{eq:global_diff}.

However, this test only supports inference for the specific evaluation set, not for new or unseen data. Treating all $d_{[jk,b,bc,l]}$ as independent observations would generally underestimate uncertainty, because it ignores dependencies within subjects, questions, or prompts; this approach reflects only the sampling variation over the observed unit-level differences. It does not account for variation that would arise from sampling new subjects, questions, or prompt variants; thus, inference is restricted to the specific evaluation set at hand. As a result, the uncertainty measured here provides a narrow view, which rarely aligns with the broader interpretation desired when analyzing benchmark results.

\paragraph{Subject Heterogeneity}
A better approach is to treat the subjects as a sample from a subject population. That is, each unit consists of a subject $t_b$, and an average score (accuracy) over all questions and all prompt variants of that subject:
\begin{equation}\label{eq:subject_diff}
    \begin{aligned}    
    d_{[jk,b,\cdot,\cdot]} = \frac{1}{N_{\mathcal{Q}(\mathcal{T})}N_{\mathcal{V}}} \sum_c\sum_l d_{[jk,b,bc,l]}.
    \end{aligned}
\end{equation}

\begin{figure}[ht]
    \centering
    \includegraphics[width=0.98\linewidth]{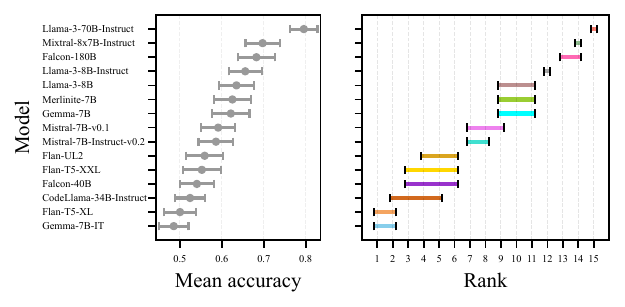}
    \caption{Scores with 95\% CIs and rank CIs from accuracy scores per subject. The overlaps between models' intervals indicate, for example, that 3 models can be ranked as the fourth from the top.}
    \label{fig:global_subject_ranking}
\end{figure}

For models $(m_j, m_k)$, we have the same t-statistic numerator $\bar{d}$ as in Equation \ref{eq:global_mu}, but here the SD quantify the spread of $d_{[jk,b,\cdot,\cdot]}$ around $\bar{d}$, and the number of units is the number of subjects $n=N_{\mathcal{T}}$. The test supports extrapolation to new subjects and their associated questions, whereas the prompt variants are considered fixed and averaged are averaged within each subject. In Figure~\ref{fig:global_subject_ranking}, we see that model rankings differ between subjects.
The hypotheses are tested with respect to ($\truescore_j^{\mathrm{subject}}, \truescore_k^{\mathrm{subject}}$), where:
\begin{equation}
    \begin{aligned}
    \truescore_j^{\mathrm{subject}}
    =
    \mathbb{E}_{T, Q(T)}\left[
        \frac{1}{N_V} 
        \sum_l
        [\mathbb{E}_S[S(m_j,T,Q(T),v_l)]
    \right].
    \end{aligned}
\end{equation}
The expectation is taken over a population of subjects ($\mathcal{T}$), questions sampled within subjects ($\mathcal{Q}(\mathcal{T})$), and the randomness of the score function ($S$).

\paragraph{Sensitivity to Prompt Variants}
A second analysis compares the average effect of each prompt across all subjects.
In this analysis, each prompt serves as the inferential unit. We define
\begin{equation}\label{eq:prompt_diff}
    \begin{aligned}
        d_{[jk,\cdot,\cdot,l]} = \frac{1}{N_{\mathcal{T}}N_{\mathcal{Q}(\mathcal{T})}} \sum_b\sum_c d_{[jk,b,bc,l]}.
    \end{aligned}
\end{equation}

The t-test is based on the variance of $\bar{d}$ (see Equation~\ref{eq:global_mu}), calculated across prompts, where the number of units $n$ equals the number of prompt variants ($N_{\mathcal{V}}$). The variance is computed across these prompt-level units and quantifies how model differences between models vary with different prompt formulations.
The estimand is:
\begin{equation}
    \begin{aligned}
     \truescore_j^{\mathrm{prompt}}
    =
    \mathbb{E}_{V}\left[
        \frac{1}{N_{\mathcal{T}}}
        \sum_b \frac{1}{N_{\mathcal{Q}(\mathcal{T})}} \sum_c
        \mathbb{E}_S[S(m_j,t_b,q_{bc},V)]
    \right].
    \end{aligned}
\end{equation}

\begin{figure}[ht]
    \centering
    \includegraphics[width=0.98\linewidth]{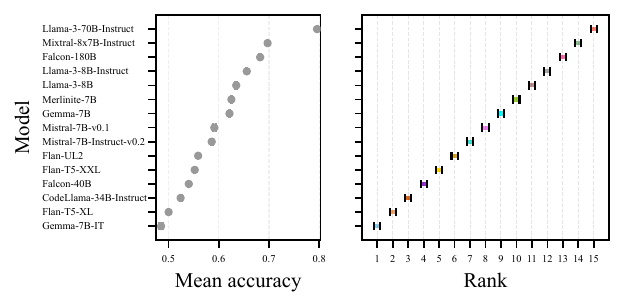}
    \caption{Scores with 95\% CIs and rank CIs from accuracy scores per prompt variant.}
    \label{fig:global_prompt_ranking}
\end{figure}

Figure~\ref{fig:global_prompt_ranking} reveals a pattern similar to Figure~\ref{fig:global_ranking_and_score}, suggesting that the number of prompts is large enough, relative to their variability, to yield stable model rankings for the observed subjects. The results are based on $N_{\mathcal{V}}=100$ prompt variants. To determine whether the number of units ($N_{\mathcal{T}}=57 < N_{\mathcal{V}}$) affects the results compared to those in Figure~\ref{fig:global_subject_ranking}, we randomly sampled $N_{\mathcal{T}}$ prompt variants from the total $N_{\mathcal{V}}$ and recalculated the rank CIs. Consistent results after subsampling suggest that subject-level variability dominates over prompt-level variability.

For the three analyses described above, the numerator of the t-statistic is identical ($\bar{d}$); the deviation in the denominator ($sd(\bar{d})$), as well as the degrees of freedom ($(n-1)$), are different.

\subsection{Subject Level Variability}
Here, we construct rank CIs for each subject independently, without regard to other subjects. We show examples of two subjects, \emph{clinical knowledge} and \emph{virology}. We change the ranking population to a specific subject at a time.
The t-test numerator for models $(m_j, m_k)$ is subject-specific, representing the average difference in model accuracy for a given subject:
\begin{equation}\label{eq:subject_mu}
\begin{aligned}
        \bar{d}^b = \score_j^b - \score_k^b = \frac{1}{N_{\mathcal{Q}(t_b)}N_{\mathcal{V}}}\sum_c\sum_l d_{[jk,b,bc,l]}.
\end{aligned}
\end{equation}

Within each subject $t_b$, we can further compare models based on their average performance, either across all questions (aggregating over prompts) or across all prompts (aggregating over questions).

For the question-based comparison within a subject, we define:
\begin{equation}\label{eq:subj_level_question_diff}
    \begin{aligned}
        d_{[jk,b,bc,\cdot]} = \frac{1}{N_{\mathcal{V}}} \sum_l d_{[jk,b,bc,l]}.
    \end{aligned}
\end{equation}
Here, paired differences are aggregated at the question level, and variance is assessed by how much $d_{[jk,b,bc,\cdot]}$ deviates from $\bar{d}^b$.

For the prompt-based comparison within a subject, we define:
\begin{equation}\label{eq:subj_level_prompt_diff}
    \begin{aligned}
        d_{[jk,b,\cdot,l]} = \frac{1}{N_{\mathcal{Q}(t_b)}} \sum_c d_{[jk,b,bc,l]}.
    \end{aligned}
\end{equation}
In this case, paired differences are aggregated at the prompt level, and the t-test variance is based on the variance of $\bar{d}^b$ calculated across prompts.

\begin{figure}[ht]
    \centering
    \begin{subfigure}[r]{0.98\linewidth}
        \centering
        \includegraphics[width=0.98\linewidth]{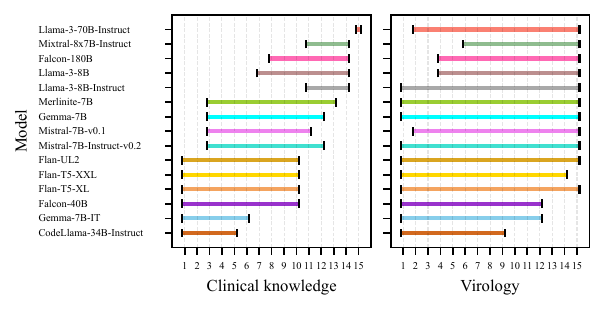}
        \caption{Rank CIs using questions as units.}
    \end{subfigure} \\
    \begin{subfigure}[r]{0.98\linewidth}
        \centering
        \includegraphics[width=0.98\linewidth]{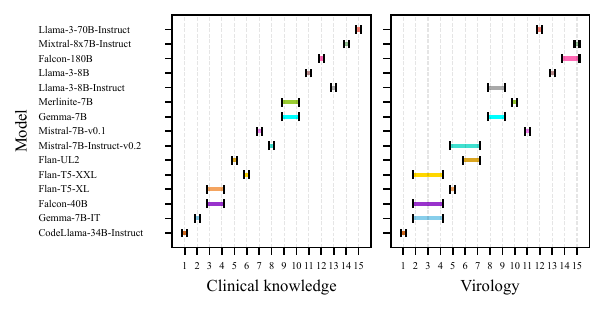}
        \caption{Rank CIs using prompt variations as units.}
    \end{subfigure}
    \caption{Subject-specific rank CIs with aggregation across questions (a) and prompts (b); the variability between questions exceed that between prompts. A model can be ranked clearly as the best for one subject (left), but not for another (right).}
    \label{fig:per_subject_ranking}
\end{figure}

Figure~\ref{fig:per_subject_ranking} presents per-subject rank CIs, calculated according to the differences defined in Equations~\ref{eq:subj_level_question_diff} and~\ref{eq:subj_level_prompt_diff}. Although many intervals overlap for both subjects, in clinical knowledge, several distinct groups of models can be identified. Both the number of questions and the number of prompt variants are 100 ($N_{\mathcal{Q}(t_b)}=N_{\mathcal{V}}$). As a result, differences between the rank CIs of two models directly reflect whether one model is statistically superior to, or equivalent with, another.

These two analyses use the same observed scores but address different inferential questions. The first analysis extrapolates model performance over questions within a fixed subject, while the second extrapolates over prompts within that subject.
\section{Hypothesis Test Specification}

In this section, we elaborate on additional specifications of the pairwise comparison process that influence the resulting rankings. We introduce the main concepts and approaches here; examples of their implementation on the MMLU benchmark can be found in Appendix~\ref{app:add_results}.

\paragraph{Statistical Difference vs Meaningful Difference}
Figure~\ref{fig:global_ranking_and_score} presents the mean accuracy and corresponding rank CIs for all models. The spread of accuracy among middle-ranked models is quite small; while the models differ significantly in the pairwise hypothesis tests, yielding rank CIs that contain only a single rank. The actual differences in performance may not be meaningful in practice. To address this, we can incorporate an indifference margin into the hypothesis specification~\cite{gibbons1979introduction}. This approach tests whether the difference between models exceeds a prespecified margin, rather than testing whether there is any difference. Formally, let $\delta > 0$ be an indifference margin, We test the following hypotheses:
\begin{equation}\label{eq:hypotheses_iz}
    \begin{aligned}
        &H_{jk;0}^{\delta}: \truescore_j - \truescore_k \leq \delta  \text{ vs } H_{jk;1}: \truescore_j - \truescore_k > \delta, \text{ and } \\
        &H_{kj;0}^{\delta}: \truescore_k - \truescore_j \leq \delta \text{ vs } H_{kj;1}: \truescore_k - \truescore_j > \delta.
    \end{aligned}
\end{equation}
When $\delta=0$, these hypotheses are identical to those defined by Equation~\ref{eq:hypotheses}. For example, in the context of the MMLU benchmark, setting $\delta = 0.02$ means that if the accuracy difference between two models is less than 2\%, they are considered practically equivalent for real-world applications; that is, such a difference is not considered meaningful in practice. Results for this example are in Appendix~\ref{app:add_indifference}.

The indifference margin can be set in advance using domain knowledge or determined empirically from the distribution of performance scores. If chosen empirically, the margin should be selected using sample splitting or alpha splitting ~\cite{hansen2000sample, huang2022simplification}.

\paragraph{Non Informative Units}

In Section~\ref{sec:mmlu}, we presented both the benchmark-level rank CIs and the per-subject rank CIs. Per-subject rank CIs can be used to improve the interpretability of benchmark rank CIs by indicating which subjects are not informative, in other words, those that do not contribute information about the differences between models. For example, in the left panel of Figure~\ref{fig:per_subject_ranking}(a), we can see that for virology, all models are almost equivalent, meaning there is no significant difference in their performance for this subject. 
To select a subset of subjects, we can split the data or the $\alpha$, construct per-subject rank CIs for all models, filter out non-informative subjects, and then construct the benchmark-level rank CIs based only on the informative ones. Moreover, identifying subjects with substantial differences between models is valuable, as this information can help select a model for a new subject. See an example in Appendix~\ref{app:add_filter}.

\section{Conclusions}

We emphasize the advantages of rank CIs over CIs for model scores, particularly for quantifying uncertainty in aggregated model performance. By exploring different ways to represent and aggregate model performance, we demonstrate how ranking based on hypothesis tests provides a clear framework for defining and measuring ranking variability and correlations between models. These definitions clarify the sources of uncertainty that are captured by the rank CIs.

In this paper, we analyzed ranking uncertainty using a single benchmark. However, LLM leaderboards typically include multiple, diverse benchmarks. In future work, we will address how to combine insights from individual benchmarks to summarize uncertainty across multiple benchmarks.

\section*{Acknowledgments}
This work was supported by the Israel Science Foundation.

\section*{Impact Statement}
Leaderboards are used by the Machine Learning community to showcase model performance. This paper shows that rankings of models within the benchmark can vary due to multiple sources of uncertainty, and that this variation should inform a more transparent and nuanced view of machine learning model performance.

\bibliography{leaderboard_ranking_workshop}
\bibliographystyle{icml2026}

\newpage
\appendix
\onecolumn
\section{Rank CIs definitions and methods}\label{app:rank_ci}

\subsection{Marginal and Simultaneous Coverage}

Our goal is to estimate the true ranks for all $N_\mathcal{M}$ models. Let $X$ be an $n \times N_\mathcal{M}$ observed aggregation of scores so that $X\sim \prob_\mu$. Here $\prob_\mu$ is a distribution family indexed by the vector of means $\mu \in R^N_\mathcal{M}$.  
Let $([L_1(X), U_1(X)], \ldots, [L_{N_\mathcal{M}}(X), U_{N_\mathcal{M}}(X)])$ denote the random data-dependent rank intervals for the $N_\mathcal{M}$ models.

\begin{definition}(Marginal Coverage)
A set of intervals has \emph{marginal coverage} at level $1 - \alpha$ if the probability that each interval cover its true rank is at least $1 - \alpha$, for any $\mu$:
\begin{equation}\label{eq:marginal_cover}
        \begin{aligned}
            \prob_\mu \big( r_j \subseteq [L_j, U_j] \big) \geq 1 - \alpha \quad \forall j \in \{1, \ldots, N_\mathcal{M}\},\quad \forall \mu \in R^{N_\mathcal{M}}.
        \end{aligned}
    \end{equation}
\end{definition}

Marginal coverage does not guarantee valid selection after ranking. In many cases, simultaneous coverage is needed. CIs with simultaneous coverage remain valid under any selection (e.g., choosing a subset of models)~\cite{benjamini2005false}, but these intervals are typically wider.

\begin{definition}(Simultaneous Coverage)
A set of intervals has \emph{simultaneous coverage} at level $1 - \alpha$ if the probability that all intervals cover their true ranks is at least $1 - \alpha$, for any $\mu$:
    \begin{equation}\label{eq:simul_cover}
        \begin{aligned}
            \prob_\mu \big( r_j \subseteq [L_j, U_j], \quad \forall j \in \{1, \ldots, N_\mathcal{M}\} \big) \geq 1 - \alpha, \quad \forall \mu \in R^{N_\mathcal{M}}. 
        \end{aligned}
    \end{equation}
\end{definition}

We use Holm's procedure to adjust p-values for multiple comparisons and control the FWER. The hypotheses are defined by the desired criterion: marginal or simultaneous rank CIs. Holm's procedure controls FWER at level $\alpha$, works efficiently, and is valid for any p-value dependence. Alternatively, other FWER-controlling methods, such as resampling algorithms that directly compute thresholds by resampling the data after removing mean differences~\cite{chetverikov2024csranks}, can be used. These can yield smaller, valid intervals when sample sizes are large enough.

\subsection{Constructing Rank CIs}

For marginal coverage, controlling the family-wise error rate (FWER) for each model individually is sufficient~\cite{holm2013confidence}.
Algorithm~\ref{alg:marginal_rank_ci} summarizes how to construct marginal rank CIs from pairwise comparisons.

\begin{algorithm}
\caption{Marginal Rank CI}\label{alg:marginal_rank_ci}
\begin{algorithmic}[1]
\STATE {\bfseries Input:}
\STATE Pairwise p-values $\{p_{jk}^b : j \neq k\}$;
\STATE FWER-controlling procedure $\phi$ (e.g., Holm), and miscoverage rate $\alpha_{tsk}$;
\FOR{$j = 1,\dots,{N_\mathcal{M}}$}
    \STATE {\bfseries Lower rank bound:}
    \STATE Apply $\phi$ to $\{p_{kj}^b : k \neq j\}$ at level $\alpha_{tsk}/2$,
           obtaining rejection vector $R_{j,L} \in \{0,1\}^{{N_\mathcal{M}}-1}$.
    \STATE $L_j^b \gets 1 + \sum_{k=1}^{{N_\mathcal{M}}-1} R_{j,L}(k)$
    \COMMENT{Count models significantly worse than $c_j$}

    \STATE {\bfseries Upper rank bound:}
    \STATE Apply $\phi$ to $\{p_{jk}^b : k \neq j\}$ at level $\alpha_{tsk}/2$,
           obtaining rejection vector $R_{j,U} \in \{0,1\}^{{N_\mathcal{M}}-1}$.
    \STATE $U_j^b \gets {N_\mathcal{M}} - \sum_{k=1}^{{N_\mathcal{M}}-1} R_{j,U}(k)$
    \COMMENT{Count models significantly better than $j$}
\ENDFOR
\STATE Return $\big\{[L_1^b, U_1^b], \ldots , [L_{N_\mathcal{M}}^b, U_{N_\mathcal{M}}^b]\big\}$
\end{algorithmic}
\end{algorithm}

\begin{theorem}\label{thm:marg}
[Holm 2013] The rank CIs $[L^b_1, U^b_1], \ldots, [L^b_{N_\mathcal{M}}, U^b_{N_\mathcal{M}}]$ constructed by Algorithm~\ref{alg:marginal_rank_ci} have marginal coverage rate $(1-\alpha)$.
\end{theorem}
See Theorem 1 in~\citet{holm2013confidence} for the formal statement and proof.
In our analyses, we use paired t-tests for significance and Holm's procedure for multiplicity control.

\begin{algorithm}
\caption{Simultaneous Rank CI}\label{alg:simultaneous_rank_ci}
\begin{algorithmic}[1]
\STATE {\bfseries Input:}
\STATE Pairwise p-values $\{p_{jk} : j \neq k\}$;
\STATE FWER-controlling procedure $\phi$ (e.g., Holm), and miscoverage rate $\alpha$;
\FOR{$j = 1,\dots,{N_\mathcal{M}}$}
    \STATE Apply $\phi$ to all p-values at level $\alpha$,
           obtaining rejection vector $R \in \{0,1\}^{{N_\mathcal{M}}({N_\mathcal{M}}-1)}$.
    
    \STATE {\bfseries Lower rank bound:}
    \STATE $L_j \gets 1 + \sum_{k=1}^{{N_\mathcal{M}}-1} R(jk)$
    \COMMENT{Count models significantly worse than $c_j$}

    \STATE {\bfseries Upper rank bound:}
    \STATE $U_j \gets {N_\mathcal{M}} - \sum_{k=1}^{{N_\mathcal{M}}-1} R(kj)$
    \COMMENT{Count models significantly better than $c_j$}
\ENDFOR
\STATE Return $\big\{[L_1, U_1], \ldots , [L_{N_\mathcal{M}}, U_{N_\mathcal{M}}]\big\}$
\end{algorithmic}
\end{algorithm}

\begin{theorem}\label{thm:simul}
[Neuhof and Benjamini 2024] The rank CIs $[L_1, U_1], \ldots, [L_{N_\mathcal{M}}, U_{N_\mathcal{M}}]$ constructed by Algorithm~\ref{alg:simultaneous_rank_ci} achieve simultaneous coverage rate $(1-\alpha)$.
\end{theorem}
See Theorem 1 in~\citet{neuhof2024confident} for details and proof.

Other methods can also construct valid rank CIs with marginal or simultaneous coverage, such as those in~\citet{al2021simultaneous, al2022simultaneous, chetverikov2024csranks, chandra2025finite, valdeira2025ranking}.
Alternative methods for constructing rank CIs require different p-value computations or more efficient FWER procedures. For matched samples, use robust alternatives to the t-test, such as the Wilcoxon signed-rank test~\cite{wilcoxon1945individual, wilcox2011introduction} (more robust, less power) or the trimmed-mean t-test~\cite{wilcox2023comparing} (intermediate power, requires sampling), when distributions are not normal; see an example in Appendix~\ref{app:add_wilcoxon}. For non-matched samples, tests like the z-test are appropriate.

\section{Additional Results}\label{app:add_results}

\subsection{Non-parametric Paired Test}\label{app:add_wilcoxon}

In our analysis, we use the paired t-test because we rank models based on accuracy, an average over binary scores, except for Figure~\ref{fig:global_ranking_and_score}. By the Central Limit Theorem, this average is approximately normally distributed.

\begin{figure}[ht]
    \centering
    \begin{subfigure}[c]{0.33\linewidth}
        \centering
        \includegraphics[width=0.98\linewidth]{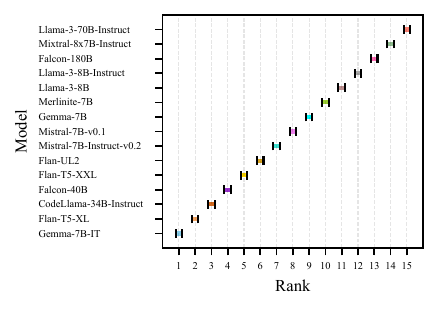}
        \caption{Rank CIs using observations as units.}
    \end{subfigure}
    \begin{subfigure}[c]{0.33\linewidth}
        \centering
        \includegraphics[width=0.98\linewidth]{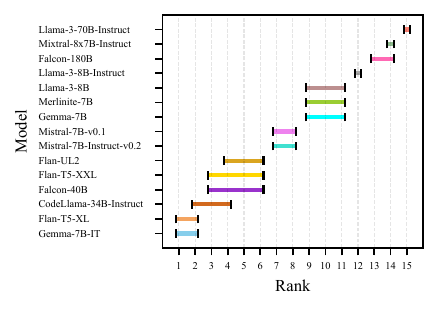}
        \caption{Rank CIs using subjects as units.}
    \end{subfigure}%
    \begin{subfigure}[c]{0.33\linewidth}
        \centering
        \includegraphics[width=0.98\linewidth]{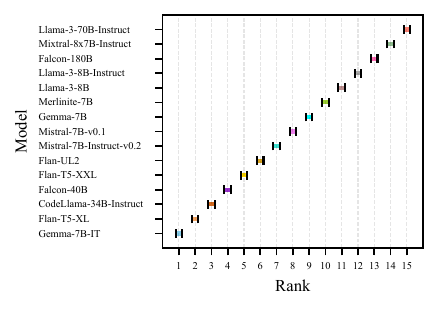}
        \caption{Rank CIs using prompt variations as units.}
    \end{subfigure}
    
    \caption{Rank CIs with Wilcoxon as the paired test and different definitions of units; observations (a), subjects (b), and prompt variants (c).}
    \label{fig:app_rank_ci_wilcoxon}
\end{figure}

However, the distribution may not be approximately normal, or even symmetric. In these cases, use a more robust test or a nonparametric test, such as the Wilcoxon signed-rank test~\cite{wilcoxon1945individual}, instead of the t-test. In Figure~\ref{fig:app_rank_ci_wilcoxon}, we show benchmark-level rank CIs using the Wilcoxon test for paired comparisons instead of the t-test, as in Figures~\ref{fig:global_ranking_and_score}, \ref{fig:global_subject_ranking}, and \ref{fig:global_prompt_ranking}. The rank CIs are the same as those obtained with the t-test, except for subjects as units (Figure~\ref{fig:global_subject_ranking}).

\subsection{Simultaneous Rank CIs}\label{app:add_simultaneous}

\begin{figure}[ht]
    \centering
    \begin{subfigure}[c]{0.48\linewidth}
        \centering
        \includegraphics[width=0.98\linewidth]{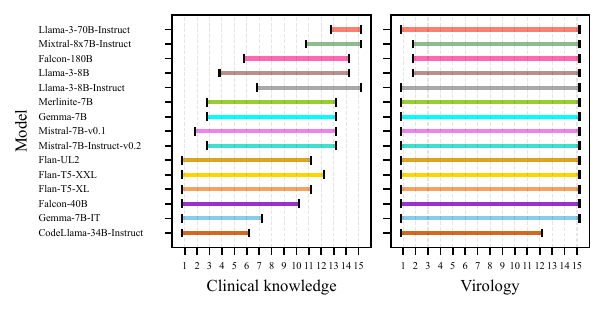}
        \caption{Rank CIs using questions as units.}
    \end{subfigure}%
    \begin{subfigure}[c]{0.48\linewidth}
        \centering
        \includegraphics[width=0.98\linewidth]{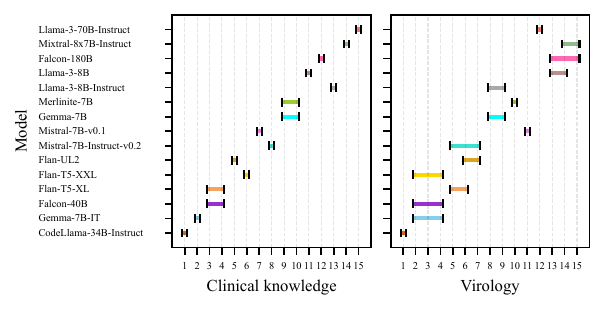}
        \caption{Rank CIs using prompt variations as units.}
    \end{subfigure}
    \caption{Subject-specific rank CIs with aggregation across questions (a) and prompts (b), with simultaneous coverage.}
    \label{fig:app_rank_ci_simultaneous}
\end{figure}

Here, we present the results of constructing simultaneous rank CIs for the same data as in Figure~\ref{fig:per_subject_ranking}. The simultaneous rank CIs in Figure~\ref{fig:app_rank_ci_simultaneous} are wider than the marginal rank CIs, as they account for multiple comparisons by considering all $N_{\mathcal{M}}(N_{\mathcal{M}}-1)$ hypotheses as a single family (see~\cite{al2021simultaneous, neuhof2024confident}). This adjustment ensures a coverage guarantee: with probability at least $1-\alpha$, the true ranks of all selected models are covered, regardless of which subset is chosen after observing the rankings. This property is especially useful when model selection is performed based on ranking outcomes.

\subsection{Indifference Zone}\label{app:add_indifference}

\begin{figure}[ht]
    \centering
    \begin{subfigure}[c]{0.33\linewidth}
        \centering
        \includegraphics[width=0.98\linewidth]{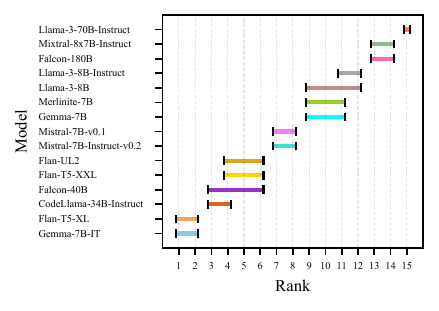}
        \caption{Rank CIs using observations as units.}
    \end{subfigure}
    \begin{subfigure}[c]{0.33\linewidth}
        \centering
        \includegraphics[width=0.98\linewidth]{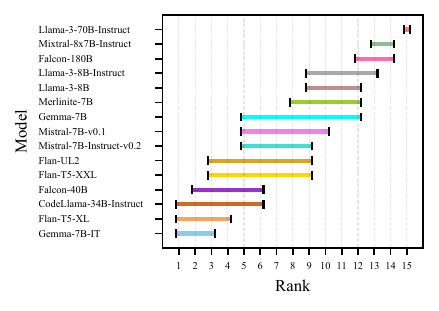}
        \caption{Rank CIs using subjects as units.}
    \end{subfigure}%
    \begin{subfigure}[c]{0.33\linewidth}
        \centering
        \includegraphics[width=0.98\linewidth]{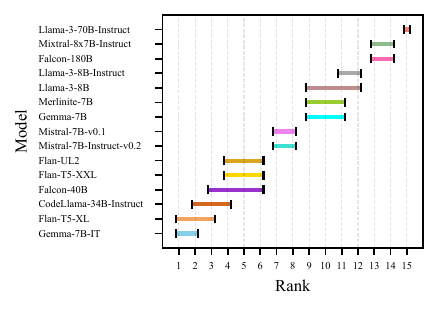}
        \caption{Rank CIs using prompt variations as units.}
    \end{subfigure}
    
    \caption{Rank CIs with indifference zone of 2\% between the accuracy of two models, and different definitions of units; observations (a), subjects (b), and prompt variants (c).}
    \label{fig:app_rank_ci_indifference}
\end{figure}

For this demonstration, we introduce an indifference margin $\delta=0.02$, representing a 2\% difference in accuracy between models. Including this margin allows us to focus on differences that are practically meaningful rather than statistically detectable. As shown in Figure~\ref{fig:app_rank_ci_indifference}, the rank intervals now overlap for all definitions of units, unlike the clear separation observed when $\delta=0$ (see Figures~\ref{fig:global_ranking_and_score} and~\ref{fig:global_prompt_ranking}). Notably, the variability between subjects remains higher than that between prompts.

\subsection{Filter Non-informative Subjects}\label{app:add_filter}

In Figure~\ref{fig:per_subject_ranking}, we observe that there is no difference between the models in virology. In this section, we analyze the distribution of per-subject rank CIs and identify non-informative subjects, those for which there is little or no distinguishable difference between models. Specifically, we define a non-informative subject as one in which 14 or 15 out of 15 models rank in the top 3, indicating that almost all models perform equivalently (see below for details). Using this criterion, we filtered out 27  subjects, including virology.

\begin{figure}[ht]
    \centering
    \begin{subfigure}[c]{0.33\linewidth}
        \centering
        \includegraphics[width=0.98\linewidth]{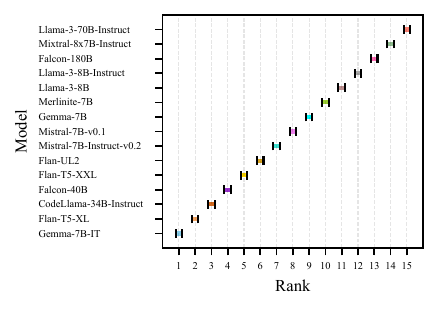}
        \caption{Rank CIs using observations as units.}
    \end{subfigure}%
    \begin{subfigure}[c]{0.33\linewidth}
        \centering
        \includegraphics[width=0.98\linewidth]{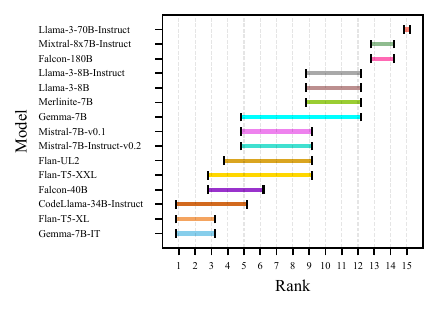}
        \caption{Rank CIs using subjects as units.}
    \end{subfigure}%
    \begin{subfigure}[c]{0.33\linewidth}
        \centering
        \includegraphics[width=0.98\linewidth]{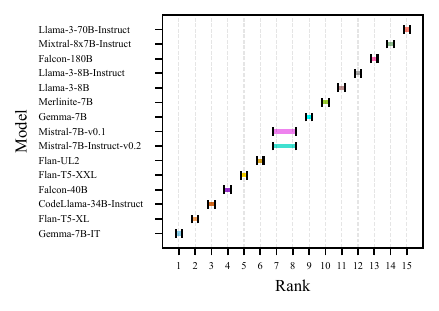}
        \caption{Rank CIs using prompt variations as units.}
    \end{subfigure}
    \caption{Rank CIs based on a subset of 30 subjects, after filtering out subjects with almost no difference between models.}
    \label{fig:app_rank_ci_filter}
\end{figure}

After filtering, we construct benchmark-level rank CIs using only the informative subjects. To maintain statistical validity, we split the miscoverage level $\alpha$ into two components: $0.3\alpha$ for constructing per-subject rank CIs and filtering, and $0.7\alpha$ for the benchmark-level rank CIs. This approach ensures appropriate control of ranking error. Figure~\ref{fig:app_rank_ci_filter} demonstrates the effect of filtering on the three sets of rank CIs: observations, subjects, and prompt variants. Filtering non-informative subjects focuses the analysis on areas where model differences are most meaningful, improving the interpretability of benchmark-level model comparisons.

\end{document}